\documentclass[11pt]{article}

\usepackage[final]{acl}

\usepackage{times}
\usepackage{latexsym}

\usepackage{amssymb}
\usepackage{amsmath}
\usepackage{booktabs}
\usepackage{multirow}
\usepackage[table]{xcolor}
\usepackage[most]{tcolorbox}

\usepackage[T1]{fontenc}

\usepackage[utf8]{inputenc}

\usepackage{microtype}

\usepackage{inconsolata}

\usepackage{graphicx}

\title{Sep-Latent: Separator Tokens as Natural Compression Anchors \\for Implicit Reasoning}

\author{
Yinlong Xu$^{1,2}$, Jianfeng Huang$^{2}$, Zhengdong Zhou$^{2}$, Mingjie Lin$^{2}$\\
\textbf{Fuyu Lv$^{2}$}, \textbf{Dan Ou$^{2}$}, \textbf{Hongxia Xu$^{1}$\thanks{\, Corresponding authors.}}, \textbf{Jian Wu$^{1}$\footnotemark[1]}\\
$^{1}$ College of Computer Science and Technology, Zhejiang University, Hangzhou, China\\
$^{2}$ Alibaba Group, Hangzhou, China\\
\texttt{\{xuyinlong,Einstein,Wujian2000\}@zju.edu.cn},
\texttt{\{lvfuyu.lfy,danou.do\}@taobao.com}\ \ \ 
\\ 
\texttt{
\{huangjianfeng.hjf,zhouzhengdong.zzd,linmingjie.lmj\}@taobao.com}\\
}

\begin{document}
\maketitle
\begin{abstract}
Large Language Models (LLMs) have demonstrated remarkable success in complex reasoning through explicit Chain-of-Thought (CoT), but the associated computational overhead and token verbosity remain significant barriers to efficiency. While implicit reasoning in continuous latent spaces offers a promising alternative, existing methods like Coconut and CODI often rely on fixed-length latent sequences and suffer from representational collapse. In this paper, we propose \textbf{Sep-Latent}, an adaptive implicit reasoning framework that utilizes separator tokens as structural semantic anchors. By introducing an Attention-Decay mechanism, we force the model to aggregate and compress reasoning logic into these anchors during training. We then align explicit and implicit pathways through multi-level distillation across all layers to ensure logical consistency. Crucially, \textbf{Sep-Latent} enables dynamic reasoning depth by adaptively terminating the latent thinking phase based on query complexity. Experimental results on the GSM8k-Aug benchmark show that Sep-Latent achieves an accuracy of 56.8\% with an average of only 5.4 tokens, significantly outperforming prior fixed-length implicit models in efficiency while narrowing the performance gap with explicit CoT. Further analysis via PCA and symbolic variant testing confirms that our method ensures semantic grounding and exhibits superior robustness to perturbations.
\end{abstract}

\section{Introduction}
Large Language Models (LLMs) have achieved remarkable success in complex reasoning tasks \cite{touvron2023llama, hurst2024gpt, shao2024deepseekmath, yang2025qwen3} through explicit Chain-of-Thought (CoT)~\cite{wei2022chain} prompting. By generating intermediate textual steps, these models can decompose intricate problems into manageable logical sequences~\cite{sun2023survey}. However, explicit CoT is often bottlenecked by significant inference latency and computational overhead due to the generation of lengthy reasoning traces~\cite{yao2022react, xu2025reason}. Furthermore, restricting reasoning to a discrete vocabulary space may lead to "over-thinking" and limits the model's ability to utilize more flexible, continuous representations for internal computation~\cite{boix2023can}.


To mitigate these issues, recent research has explored implicit reasoning~\cite{deng2405explicit, pfau2024let, goyal2023think}, where the reasoning process is performed within the model's continuous latent space. Pioneer works like Coconut~\cite{hao2022training} and CODI~\cite{shen2025codi} demonstrate that CoT trajectories can be compressed into compact latent representations through distillation or curriculum learning. More recently, SIM-CoT~\cite{wei2025sim} introduced an extra supervised LLM to align implicit states with explicit reasoning steps to ensure logical consistency. Despite these advances, existing methods often utilize a fixed number of latent tokens, which lacks the flexibility to adapt to varying problem complexities and risks "semantic homogenization"—where latent states collapse into uninformative noise.

\begin{figure*}[t]
    \centering
    \includegraphics[width=\linewidth]{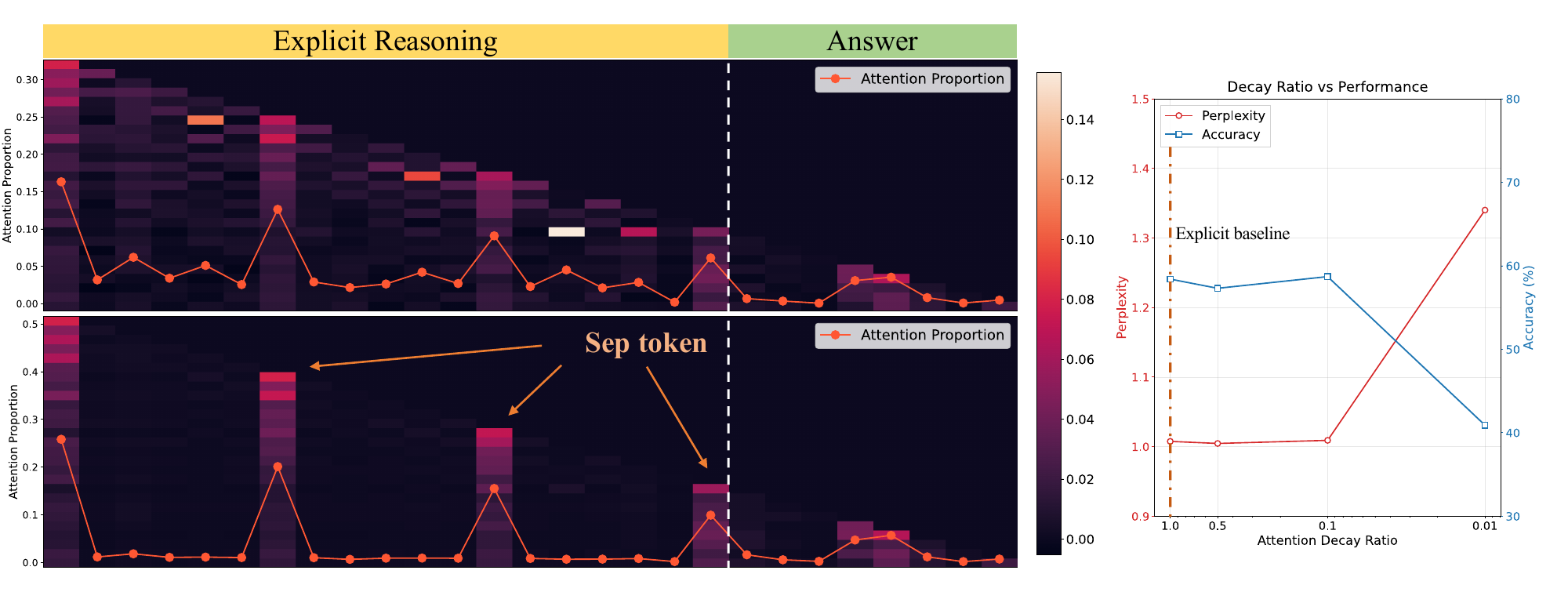}
    \caption{Language models naturally allocate strong attention to separator tokens during reasoning. Increasing attention concentration on separator tokens preserves reasoning accuracy and perplexity with negligible performance variation}
    \label{combined_attention}
\end{figure*}


Inspired by the observation that LLMs naturally prioritize separator tokens in \ref{combined_attention}, we amplify attention concentration on these tokens through an attention-decay mechanism, enabling compact reasoning compression with negligible changes in perplexity and accuracy. We propose \textbf{Sep-Latent}, a framework utilizing Separator-guided semantic anchors and a progressive Attention-Decay mechanism. By selectively suppressing attention weights of non-separator tokens during training, we force the model to compress essential reasoning logic into structural tokens (e.g., ",","."). A dual-path distillation objective then aligns these compressed explicit representations with the implicit student path at both token and sequence levels.Our method introduces three significant improvements over existing frameworks:


\textbf{Attention-Decay Mechanism:}
We propose an attention-decay mechanism that progressively compresses intermediate reasoning information into separator tokens, enabling efficient latent reasoning while preserving long-range reasoning coherence.

\textbf{Online-Policy Distillation:}
We introduce an online-policy distillation strategy where the model continuously distills its own on-policy latent reasoning trajectories, improving reasoning stability while significantly reducing training memory overhead through self-distillation on compact latent states.

\textbf{Dynamic Implicit Exit:}
Sep-Latent enables adaptive termination. By applying an supervision on an exit token, the model dynamically adjusts reasoning depth based on query complexity and current reasoning states.

Experimental results across multiple benchmarks, including math benchmark and commonsense benchmarks, 
demonstrate that Sep-Latent significantly outperforms previous implicit methods. On average, our method achieves a competitive accuracy of 56.8\% on GSM8k~\cite{cobbe2021training} using only 5.4 tokens, representing a more than two-fold increase in token efficiency compared to CODI and SIM-CODI. Furthermore, PCA visualizations confirm that our latent tokens are semantically diverse and well-grounded in the vocabulary space, effectively avoiding representational collapse. These findings establish Sep-Latent as a highly efficient and robust framework for internalized LLM reasoning.

\section{Related Work}
\subsection{Implicit Reasoning}
In recent years, researchers have increasingly focused on implicit reasoning, motivated by the fact that explicit CoT can substantially increase inference cost in long-sequence settings and often produces redundant intermediate steps~\cite{zhang2025soft,li2025implicit}. To address this, a growing body of work enables LLMs to perform reasoning in a continuous latent space, thereby avoiding the need to explicitly generate CoT text~\cite{strobl2024formal,merrill2024expressive}.

To preserve reasoning capability without generating CoT, existing methods largely follow two directions: representation alignment and latent-trajectory learning. ICoT~\cite{deng2023implicit} promotes the internalization of reasoning by aligning hidden states between a teacher model that generates CoT and a student model that outputs answers directly, using intermediate-representation supervision. Stepwise Internalization~\cite{deng2405explicit} further proposes a staged fine-tuning procedure that gradually removes intermediate reasoning supervision, encouraging the model to internalize the reasoning process. In parallel, Coconut~\cite{hao2024training} is among the first to distill explicit reasoning traces into a continuous latent space to learn latent reasoning; subsequent work~\cite{xu2025softcot,sutoken} continue to explore latent-space reasoning, yet still relies on explicit CoT generation during training or inference. SIM-COT~\cite{wei2025sim} additionally introduces supervised alignment between implicit reasoning and explicit thought steps to enhance reasoning consistency and accuracy. In contrast to prior methods, we introduce an attention-decay mechanism that organizes reasoning information in latent space through hierarchical structural guidance and semantic compression constraints, thereby enabling structured, controllable, and efficient implicit reasoning.

\subsection{Online-Policy Knowledge Distillation}
Knowledge distillation (KD)\cite{gou2021knowledge,xu2024survey} compresses models by transferring behavioral or representational knowledge from a teacher to a student, and has recently expanded from output-level distillation to distilling the reasoning process itself\cite{hsieh2023distilling,ho2023large}.
System~2 Distillation~\cite{yu2024distilling} proposes a self-distillation framework in which the same model alternates between teacher and student roles across stages and converts deep, explicit System-2 reasoning such as Tree-of-Thought or CoT into fast, heuristic System-1 reasoning. CODI~\cite{shen2025codi} compresses complex CoT trajectories into compact latent representations via iterative self-distillation, thereby enabling implicit reasoning. UnDIAL~\cite{dong2025undial} further employs self-distillation to preserve the model's original behavioral characteristics. TwT~\cite{xu2025twt} introduces a three-stage habitual-reasoning distillation framework that uses multi-teacher guidance and dual-criterion sampling and compresses explicit reasoning into latent capabilities. Building on these advances, we propose a joint reasoning optimization framework that jointly trains the model in explicit and implicit reasoning modes and enforces semantic consistency between them under structured supervision, which enables smooth knowledge transfer across the two reasoning pathways.

\section{Method}

\begin{figure*}[ht]
    \centering
    \includegraphics[width=\textwidth]{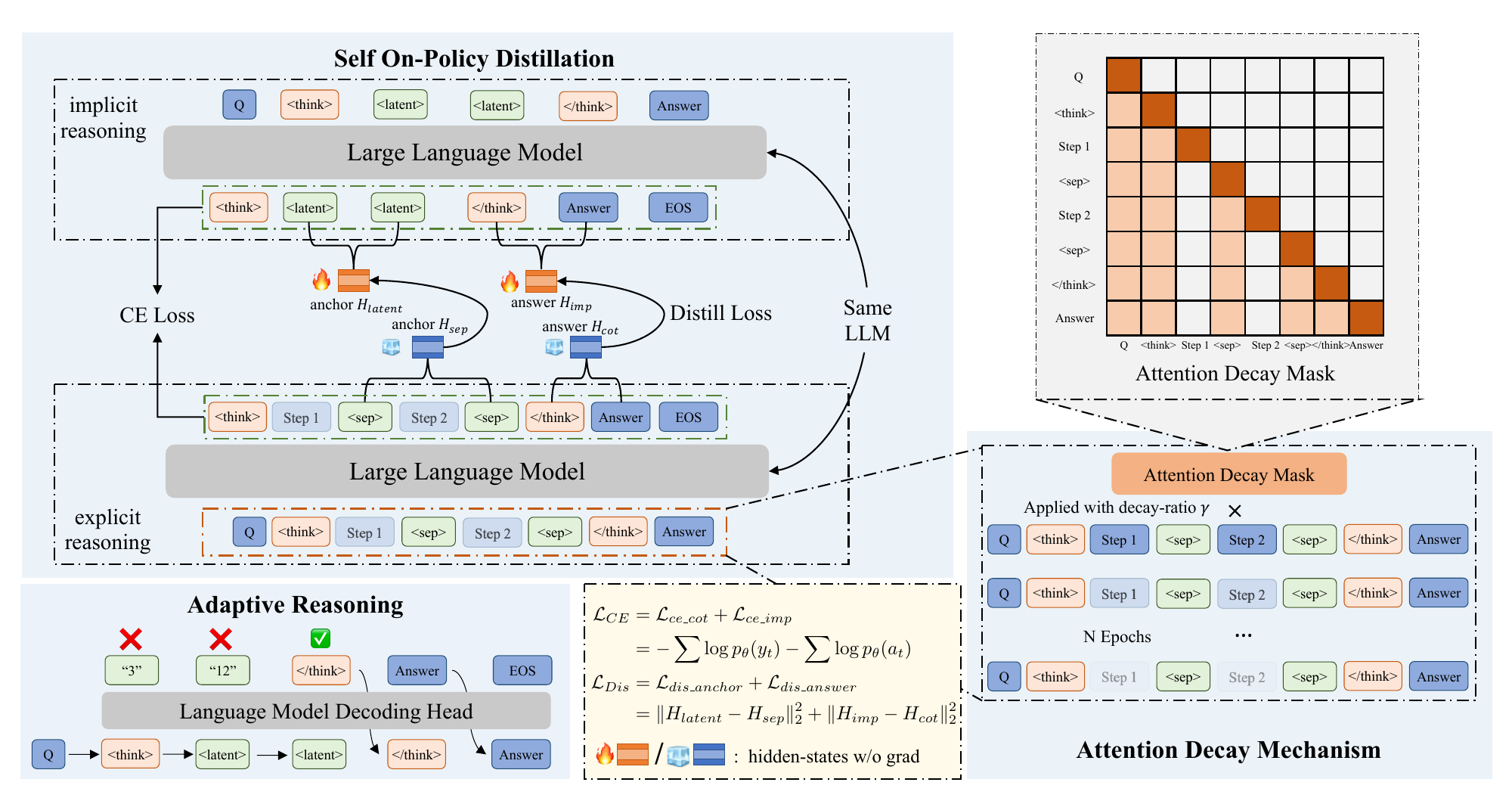}
    \caption{Algorithm of Sep-Latent: As the training epochs proceed, LLM pays less and less attention to the reasoning step, and the information is gradually compressed into <sep>. It should be noted that although the figure uses two LLMS to illustrate the algorithm, there is actually one model during the actual training process.}
    \label{algorithm-structure}
\end{figure*}

As shown in Fig.\ref{algorithm-structure}, we propose a multi-stage training curriculum that transitions from Explicit Chain-of-Thought~(CoT) to Implicit Reasoning by utilizing Separator Tokens~(e.g. "\textbackslash n", ",", ".") as semantic anchors. The core of this method involves a progressive Attention-Decay mechanism that encourages the model to compress semantic information from reasoning steps into these anchor tokens.

\subsection{Notation}
\label{Notation}

Let $\mathcal{V}$ be the size of the vocabulary, $d$ be the dim of hidden states, $e(\cdot) \in \mathbb{R}^d$ be the token embedding function with a matrix $E \in \mathbb{R}^{\lvert \mathcal{V} \rvert \times d}$, and $w(\cdot)\in \mathcal{V}$ be the language modeling head function with a matrix $W \in \mathbb{R}^{d \times \lvert \mathcal{V} \rvert}$. We define $x=(x_1, x_2, \dots, x_T)\in \mathcal{V}^T$ as the textual input query and employ our experiments on an autoregressive LLM~$p_{\theta}$ with $L$ attention layers and parameters $\theta$, the LLM~$p_{\theta}$ accept an embedded tokens sequence: 
\begin{equation*}
    U_{T}=(u_1, u_2, \dots, u_T), \quad u_i \in \mathbb{R}^d
\end{equation*}
where $u_i=e(x_i)$ in explicit reasoning, and will be projecting of the output of the last attention layer of the previous sequences in implicit reasoning. We define the output of the $k$-th attention layer with the input $U_{T}$ as: 
\begin{equation*}
    h_T^{(k)}=p_\theta^{(k)}(U_{T}), \quad h_T^{(k)} \in \mathbb{R^d}
\end{equation*}

We assume explicit reasoning chain consists of several steps $S = \{s_1, s_2, \dots, s_m\}$, the $k$-th textual step is $s_k=(y_{k,1}, y_{k,2}, \dots, y_{k,T_k})\in \mathcal{V}^{T_k}$, and the answer is $a=(a_1, a_2, \dots,a_{T_a})\in \mathcal{V}^{T_a}$.

\subsection{Explicit CoT with Attention-Decay}


To encourage the model to compress reasoning complexity into sparse semantic anchors, we introduce an Attention-Decay mechanism in the explicit path. Unlike standard CoT's uniform attention, we selectively suppress non-separator tokens by adding separators $V = \{v_1, \dots, v_{m}\}$ to the sequence $S$. The total sequence is represented as:

\begin{equation*}
    Y_{cot}=(x, v_0, s_1,\dots, s_m, v_m, a).
\end{equation*}

We define $\mathcal{I}_{steps}$ as the set of indices corresponding to the steps tokens $S$ in $Y_{cot}$. For the $l$-th attention layer, let $A^{(l)}_{i,j}$ denote the raw attention weight from query position $i$ to key position $j$. The modified attention weight $\hat{A}^{(l)}_{i,j}$ is defined as:
\begin{equation}
    \hat{A}^{(l)}_{i,j} = 
    \begin{cases} 
        A^{(l)}_{i,j} \cdot \gamma & \text{if } j \in \mathcal{I}_{\text{steps}} \\ 
        A^{(l)}_{i,j} & \text{otherwise} 
    \end{cases}
\end{equation}
where $\gamma \in [0, 1]$ is the decay factor. By reducing the attention weights of the step-specific tokens $s_k$, the model is forced to aggregate and propagate necessary reasoning information through the hidden states of the separator tokens $v_k$, effectively treating them as "summary latents".

We also define $\mathcal{I}_{sep}$ and $\mathcal{I}_{cot\_ans}$ as the set of indices corresponding to the separator tokens $V$ and answer tokens in $Y_{cot}$, we keep the hidden states of separator tokens and answer tokens of all layers in $p_\theta$ for subsequent alignment with implicit inference at each forward process:
\begin{equation}
\begin{alignedat}{2}
    H_{\text{sep}} &= \sum_{i \in \mathcal{I}_{\text{sep}}} &&\sum_{k \leq L} p_{\theta}^{k}(U_i) \\
    H_{\text{cot}} &= \sum_{i \in \mathcal{I}_{\text{cot\_ans}}} &&\sum_{k \leq L} p_{\theta}^{k}(U_i)
\end{alignedat}
\end{equation}

\subsection{Implicit Reasoning Alignment}
In the implicit reasoning path, the explicit textual steps $s_k$ are replaced by a sequence of continuous latent tokens $Z = \{z_1, \dots, z_m\}$. The resulting sequence is represented as:
\begin{equation*}
    Y_{\text{imp}} = (x, z_1, z_2,\dots, z_m, a).
\end{equation*}

Unlike the explicit phase where inputs are derived from the embedding function $e(\cdot)$, the input vector $u_t$ for a latent position $t \in \mathcal{I}_{\text{latent}}$ is generated through a projection of the model's internal state. Specifically, we employ a multi-layer perceptron (MLP) to project the last-layer hidden state of the previous token into the embedding space of the current step:
\begin{equation}
    \label{implicit-reason}
    u_t = \text{MLP}(h_{t-1}^{(L)}), \quad h_{t-1}^{(L)} \in \mathbb{R}^d
\end{equation}
where $h_{t-1}^{(L)}$ is the output of the $L$-th attention layer at position $t-1$. This recursive construction enables the implicit path to bypass vocabulary constraints while maintaining autoregressive dependencies.

Similar to the explicit phase, we define $\mathcal{I}_{latent}$ and $\mathcal{I}_{\text{imp\_ans}}$ as the indices for separator and answer tokens within $Y_{\text{imp}}$. We aggregate the hidden states across all $L$ layers to capture the hierarchical reasoning features:
\begin{equation}
\begin{alignedat}{2}
    H_{\text{latent}}   &= \sum_{i \in \mathcal{I}_{\text{latent}}}   && \sum_{k \leq L} p_{\theta}^{k}(U_i) \\
    H_{\text{imp}} &= \sum_{i \in \mathcal{I}_{\text{imp\_ans}}} && \sum_{k \leq L} p_{\theta}^{k}(U_i)
\end{alignedat}
\end{equation}

These representations serve for alignment with the "compressed" knowledge generated in the explicit path under the attention-decay mechanism.

\subsection{Training Objectives}
\textbf{Cross-Entropy Supervision.} To maintain the model's fundamental capability, we apply cross-entropy (CE) losses to both paths. The probability of generating the $t$-th token is derived from the last-layer hidden state $h_{t-1}^{(L)}$ as: 
\begin{equation}
    p_\theta(y_t | Y_{<t}) = \text{Softmax}(W h_{t-1}^{(L)})_{y_t},
\end{equation}
where $W$ is matrix of the LM-head mentioned in \ref{Notation}. For the explicit path, we supervise the entire sequence $Y_{cot}$ (including reasoning steps and the answer) and for the implicit path, supervision is applied only to the final answer$a$:
\begin{equation}
\begin{alignedat}{2}
    \mathcal{L}_{\text{ce\_cot}} &= -\sum_{y_t \in {S, V, a}} \log p_\theta(y_t \mid Y_{cot,<t}) \\
    \mathcal{L}_{\text{ce\_imp}} &= - \sum_{t=1}^{T_a} \log p_\theta(a_t | x,z_{1:m},a_{<t})
\end{alignedat}
\end{equation}

\textbf{Token-level Distillation.} We perform distillation at both the structural anchors and the final answer positions to align two paths. By minimizing the squared Euclidean distance between the aggregated hidden states $H$, we force the implicit path to mimic the "compressed" semantic space of the explicit path:
\begin{equation}
\begin{aligned}
    \mathcal{L}_{\text{dis\_anchor}} &= \lvert\lvert H_{\text{latent}} - H_{\text{sep}} \rvert\rvert_2^2 \\
    \mathcal{L}_{\text{dis\_answer}} &= \lvert\lvert H_{\text{imp}} - H_{\text{cot}} \rvert\rvert_2^2
\end{aligned}
\end{equation}
$\mathcal{L}_{\text{dist\_sep}}$ ensures that the implicit latents and separators encapsulate the intermediate logic, while $\mathcal{L}_{\text{dist\_ans}}$ aligns the terminal hidden states to ensure consistency in the final prediction.

\textbf{Training Objective.} The final training loss $\mathcal{L}$ is a weighted sum of the linguistic grounding and representation alignment tasks:
\begin{equation}
\begin{aligned}
    \mathcal{L}_{CE} &= \mathcal{L}_{\text{ce\_cot}} + \mathcal{L}_{\text{ce\_imp}}\\
    \mathcal{L}_{Dis} &= \mathcal{L}_{\text{dis\_anchor}} + \mathcal{L}_{\text{dis\_answer}}\\
    \mathcal{L} &= \mathcal{L}_{CE}+\mathcal{L}_{Dis}
\end{aligned}
\end{equation}

It should be noted that explicit reasoning is not required in the inference phase, and the inference behaviors are demonstrated in the appendix~\ref{inference-mode}.

\section{Experiments}
\label{Experiments}

\renewcommand{\arraystretch}{0.7}

\begin{table*}[ht]
    \centering
    \small
    \setlength{\tabcolsep}{4.5pt}
    \caption{Main results on Llama3.2-1B-Instruct and Qwen3-0.6B. We report accuracy (\%) and token consumption on each benchmark. For Sep-Latent, we report the best performance with the $decay\_ratio = 0.25$. Bold indicates the best result, while underlined values denote the second-best result.}
    \label{1b-result}
    \begin{tabular}{l|cc|cc|cc|cc|cc}
        \toprule
        \multirow{2}[2]{*}{\textbf{Method}} & \multicolumn{2}{c|}{GSM8k-Aug} 
        & \multicolumn{2}{c|}{GSM-Hard} & \multicolumn{2}{c|}{MultiArith} & \multicolumn{2}{c|}{SVAMP} & \multicolumn{2}{c}{Avg.} \\
        \cmidrule(lr){2-3} \cmidrule(lr){4-5} \cmidrule(lr){6-7} \cmidrule(lr){8-9} \cmidrule(lr){10-11}
         & Acc. (\%) & tokens & Acc. (\%) & tokens & Acc. (\%) & tokens & Acc. (\%) & tokens & Acc. (\%) & tokens \\
        \midrule
        \multicolumn{11}{c}{\textbf{Llama3.2-1B}} \\
        \midrule
        SFT-CoT     & \underline{56.2} & 25.3 & \textbf{13.9} & 23.1 & 96.7 & 11.1 & \textbf{65.7} & 12.1 & 58.1 & 17.9 \\
        SFT-Direct  & 28.8 & 1.2  & 6.3  & 1.9  & 50.3 & 1.9  & 26.7 & 1.9  & 28.0 & 1.9  \\
        \midrule
        $\text{Coconut}$        & 33.2 & 7.2 & 7.0  & 5.9 & 63.3 & 5.9 & 43.7 & 5.9 & 36.8 & 5.9 \\
        $\text{SIM-Coconut}$    & 42.2 & 7.2 & 9.3  & 5.9 & 87.7 & 5.9 & 43.9 & 5.9 & 45.8 & 5.9 \\
        \midrule
        CODI        & 50.7 & 7.2 & 11.7 & 8.2 & 97.0 & 7.0 & 61.7 & 7.1 & 55.3 & 7.4 \\
        SIM-CODI    & 52.3 & 7.2 & 11.8 & 8.3 & \textbf{97.8} & 7.0 & 61.3 & 7.1 & 55.8 & 7.4 \\
        \midrule
        Sep-Latent(ours)& \textbf{56.8} & 5.4 & \underline{13.1} & 5.9 & \underline{97.7} & 5.9 & \underline{64.0} & 5.9 & 57.9 & 5.9 \\
        w/o Decay       & 50.7 & 5.4 & 11.2 & 5.9 & 96.0 & 5.9 & 60.5 & 5.9 & 54.6 & 5.9 \\
        w/o Distill     & 47.2 & 5.4 & 10.0 & 5.9 & 82.0 & 5.9 & 52.0 & 5.9 & 47.8 & 5.9 \\
        w/o Decay \& Distill & 37.6 & 5.4 & 6.8  & 5.9 & 55.0 & 5.9 & 35.0 & 5.9 & 33.6 & 5.9 \\
        \midrule
        \multicolumn{11}{c}{\textbf{Qwen3-0.6B}} \\
        \midrule
        SFT-CoT     & \underline{52.3} & 22.5 & \textbf{12.5} & 23.1 & 89.2 & 12.1 & \textbf{60.1} & 13.1 & 53.5 & 17.9 \\
        SFT-Direct  & 25.1 & 1.2  & 5.6  & 1.9  & 45.6 & 1.9  & 23.5 & 1.9  & 24.9 & 1.9  \\
        CODI        & 46.2 & 7.2 & 10.5 & 8.2 & 89.5 & 7.0 & 56.0 & 7.1 & 50.6 & 7.4 \\
        SIM-CODI    & 47.8 & 7.2 & 10.6 & 8.3 & \underline{90.2} & 7.0 & 55.8 & 7.1 & 51.1 & 7.4 \\
        Sep-Latent(ours)& \textbf{53.5} & 5.4 & \underline{11.8} & 5.9 & \textbf{90.5} & 5.9 & \underline{58.5} & 5.9 & 53.6 & 5.9 \\
        \bottomrule
    \end{tabular}
\end{table*}

\begin{figure*}[ht]
    \centering
    \includegraphics[width=\textwidth]{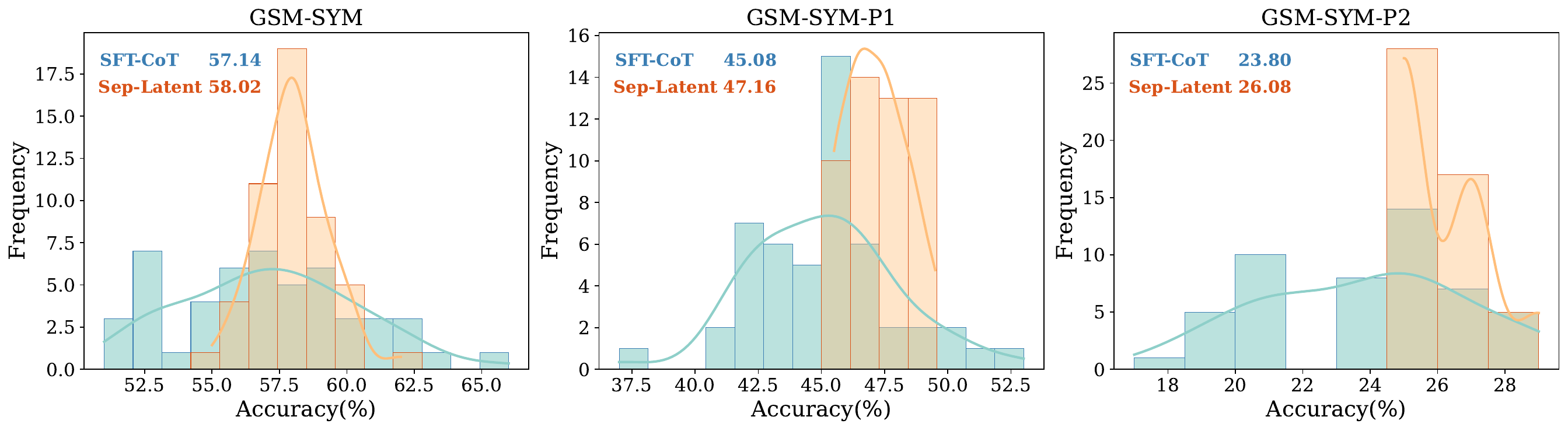}
    \caption{The result of SFT-CoT and Sep-Latent on GSM-SYM benchmarks.}
    \label{sym-results}
\end{figure*}

\subsection{Training Datasets}
Recent studies indicate that fine-tuning models with the GSM8k-Aug~\cite{deng2023implicit} dataset can significantly enhance the model's reasoning capabilities in mathematical tasks. GSM8k-Aug is an augmented dataset derived from the original GSM8k~\cite{cobbe2021training}, expanded to approximately 385k examples with GPT-4.

The dataset is processed to remove natural language reasoning chains, 
intending to force the model to internalize the intermediate reasoning process into specific mathematical calculations rather than discrete textual tokens. We present examples of the data used for explicit and implicit reasoning training in the appendix~\ref{data-example}.

\subsection{Evaluation Benchmarks}


To evaluate the robustness and generalization of Sep-Latent, we conduct comprehensive evaluations across multiple benchmark categories: GSM8k~\cite{cobbe2021training} serves as the in-domain test set, while MultiArith, SVAMP~\cite{patel2021nlp}, and GSM-Hard~\cite{gao2023pal} are used to assess out-of-domain mathematical reasoning under structural and numerical distribution shifts. We further evaluate logical and commonsense reasoning capabilities using LogiQA~\cite{logiqa}, CSQA~\cite{commonqa}, and StrategyQA~\cite{geva2021did}. Finally, we include GSM-SYM~\cite{mirzadeh2024gsm} to test the stability of our framework against symbolic perturbations. Detailed dataset distributions are provided in Appendix~\ref{benchmark-detail}.

\subsection{Baselines}
To provide a comprehensive comparison, we evaluate our method against the following representative baselines:(1)~SFT-COT: Supervised fine-tuning (SFT) on CoT-annotated data, LLMs will output an explicit reasoning chain before outputing the answer.(2)~SFT-Direct: Supervised fine-tuning (SFT) only on the answer, LLMs will direct output the answer without any thinking. (3)~Coconut~\cite{hao2022training}: By curriculum learning, LLMs gradually replace textual steps by latent expression. (4)~CODI~\cite{shen2025codi}: By using knowledge distillation to align the hidden states of the answer of implict LLMs to explicit LLMs. (5)~SIM-COT~\cite{wei2025sim}: An auxiliary model is introduced to decode the latent space reasoning of the original method so as to supervise the intermediate process. We follow the training setup of the previous and employ our experiments on Llama3-1B/3B/8B~\cite{dubey2024llama}, 
the detailed baselines setting are provided in Appendix~\ref{implementation-details}.

\section{Main Results}
\subsection{In-Domain Mathematical Reasoning}
We report the result of Llama3.2-1B and Qwen3-0.6B in Table~\ref{1b-result}, in terms of in-domain performance on the GSM8k-Aug test set, Sep-Latent demonstrates a significant competitive advantage over existing implicit reasoning frameworks. Our method achieves an accuracy of 56.8\%, marking a substantial improvement of +4.5\% over the current state-of-the-art distillation method, SIM-CODI (52.3\%). When compared against SIM-CODI (52.3\%), which incorporates advanced step-level supervision, Sep-Latent maintains a performance edge. Most importantly, our approach effectively bridges the performance gap between implicit and explicit reasoning, nearing the accuracy of the SFT-CoT baseline (56.2\%). 
These results demonstrate that Sep-Latent outperforms SIM-based frameworks while eliminating the need for auxiliary decoding models, thereby significantly reducing memory overhead.

\begin{table*}[htbp]
    \centering
    \small
    \setlength{\tabcolsep}{4.5pt}
    \caption{Results on Llama3.2-3B and Llama3.1-8B models. We report accuracy (\%) and token consumption on each benchmark. Bold indicates the best result, while underlined values denote the second-best result.}
    \label{big-llm}
    \begin{tabular}{l|cc|cc|cc|cc|cc}
        \toprule
        \multirow{2}[2]{*}{\textbf{Method}} & \multicolumn{2}{c|}{GSM8k-Aug} 
        & \multicolumn{2}{c|}{GSM-Hard} & \multicolumn{2}{c|}{MultiArith} & \multicolumn{2}{c|}{SVAMP} & \multicolumn{2}{c}{Avg.} \\
        \cmidrule(lr){2-3} \cmidrule(lr){4-5} \cmidrule(lr){6-7} \cmidrule(lr){8-9} \cmidrule(lr){10-11}
         & Acc. (\%) & tokens & Acc. (\%) & tokens & Acc. (\%) & tokens & Acc. (\%) & tokens & Acc. (\%) & tokens \\
        \midrule
        \multicolumn{11}{c}{\textbf{Llama3.2-3B}} \\
        \midrule
        SFT-CoT & \textbf{71.5} & 27.7 & \textbf{17.0} & 22.4 & 98.3 & 11.4 & 71.0 & 12.4 & 64.5 & 18.5 \\
        SFT-Direct & 38.3 & 1.2 & 9.5 & 1.4 & 88.7 & 1.4 & 52.9 & 1.4 & 47.4 & 1.4 \\

        CODI & 60.8 & 8.2 & 14.3 & 11.5 & 98.7 & 11.5 & \underline{73.3} & 11.5 & 61.8 & 10.7 \\
        SIM-CODI & 62.3 & 8.2 & 14.6 & 11.5 & \textbf{98.8} & 11.5 & 74.9 & 11.5 & 62.7 & 10.7 \\

        Sep-Latent(ours) & \underline{64.7} & 5.5 & \underline{16.3} & 5.7 & \underline{98.5} & 5.7 & \textbf{75.1} & 5.7 & 63.7 & 5.7 \\
        \midrule
        \multicolumn{11}{c}{\textbf{Llama3.1-8B}} \\
        \midrule
        SFT-CoT & \textbf{71.7} & 27.4 & \textbf{16.5} & 22.2 & 98.3 & 13.2 & 73.1 & 12.2 & 64.9 & 18.8 \\
        SFT-Direct & 39.5 & 1.2 & 9.8 & 1.6 & 88.0 & 1.6 & 55.3 & 1.6 & 48.2 & 1.5 \\
        CODI & 61.1 & 8.2 & 15.5 & 11.5 & \underline{99.5} & 11.5 & 78.1 & 11.5 & 63.6 & 10.7 \\
        SIM-CODI & 64.1 & 8.2 & 16.1 & 11.5 & \textbf{100.0} & 11.5 & \textbf{78.7} & 11.5 & 64.7 & 10.7 \\
        Sep-Latent(ours) & \underline{68.8} & 5.5 & \underline{16.3} & 5.7 & 98.7 & 5.7 & \underline{73.3} & 5.7 & 64.3 & 5.7 \\
        \bottomrule
    \end{tabular}
\end{table*}

\subsection{Out-of-Domain Reasoning}

\textbf{Math problems of different difficulties.}Results in Table~\ref{1b-result} underscore the superior generalization of Sep-Latent. This advantage is particularly evident in challenging scenarios like GSM-Hard, where Sep-Latent reaches 13.1\% accuracy. On SVAMP, our model shows heightened resilience to structural perturbations with 62.0\% accuracy, surpassing both baselines. Furthermore, on MultiArith, Sep-Latent maintains a high performance of 95.7\%, nearly matching the explicit SFT-CoT teacher (96.7\%) while remaining significantly more efficient.

\textbf{Variant Mathematical Reasoning.}
Traditional explicit reasoning are proved to be fragile when facing variant mathematical, this could be due to data leakage during the pre-training phase or a misplaced focus on irrelevant conditions. As illustrated in Figure~\ref{sym-results}, Sep-Latent consistently demonstrates superior robustness compared to the explicit SFT-CoT baseline across all symbolic variant benchmarks. By shifting computation into the continuous latent space and bypassing the need for deterministic textual traces, Sep-Latent effectively mitigates the performance degradation typically caused by superficial perturbations in problem statements. These results suggest that implicit reasoning fosters a more generalized understanding of mathematical logic, making the model less susceptible to the linguistic biases or irrelevant textual cues that often plague explicit chain-of-thought methods.

\renewcommand{\arraystretch}{0.7}
\begin{table}[htbp]
    \centering
    \setlength{\tabcolsep}{4pt}
    \caption{Commonsense reasoning results on Llama3.2-1B-Instruct. We report accuracy(\%) on three datasets: LogiQA, CommonsenseQA and StrategyQA.}
    \label{commonsense-result}
    \begin{tabular}{lcccc}
        \toprule
         \textbf{Method} & \textbf{Logi} & \textbf{CSQA} & \textbf{Strategy} & \textbf{Avg.}\\
         \midrule
         SFT-CoT &39.8 & 68.2 & 41.1 & 49.7\\
         SFT-Direct & 46.2 & 74.9 & 63.4 & 61.5\\
         \midrule
         CODI & 44.6 & 72.4& 61.8 & 59.6\\
         SIM-CODI & 45.2 & 73.2& 63.1 & 60.5\\
         \midrule
         Sep-Latent & 47.1 & 73.9 & 62.9 & 61.3\\
        \bottomrule
    \end{tabular}
\end{table}


\textbf{Commonsense Reasoning.} We further demonstrate the effectiveness of Sep-Latent across logical and linguistic tasks. As shown in Table~\ref{commonsense-result}, Sep-Latent achieves a 61.3\% average accuracy, outperforming both CODI (59.6\%) and SIM-CODI (60.5\%). Our method nearly matches the SFT-Direct performance ceiling (61.5\%) while significantly surpassing the verbose SFT-CoT baseline (49.7\%). Notably, on the LogiQA benchmark, Sep-Latent attains the highest accuracy of 47.1\%, demonstrating its superior capacity for logical reasoning compared to all other evaluated methods.


\begin{figure*}[ht]
    \centering
    \includegraphics[width=\textwidth]{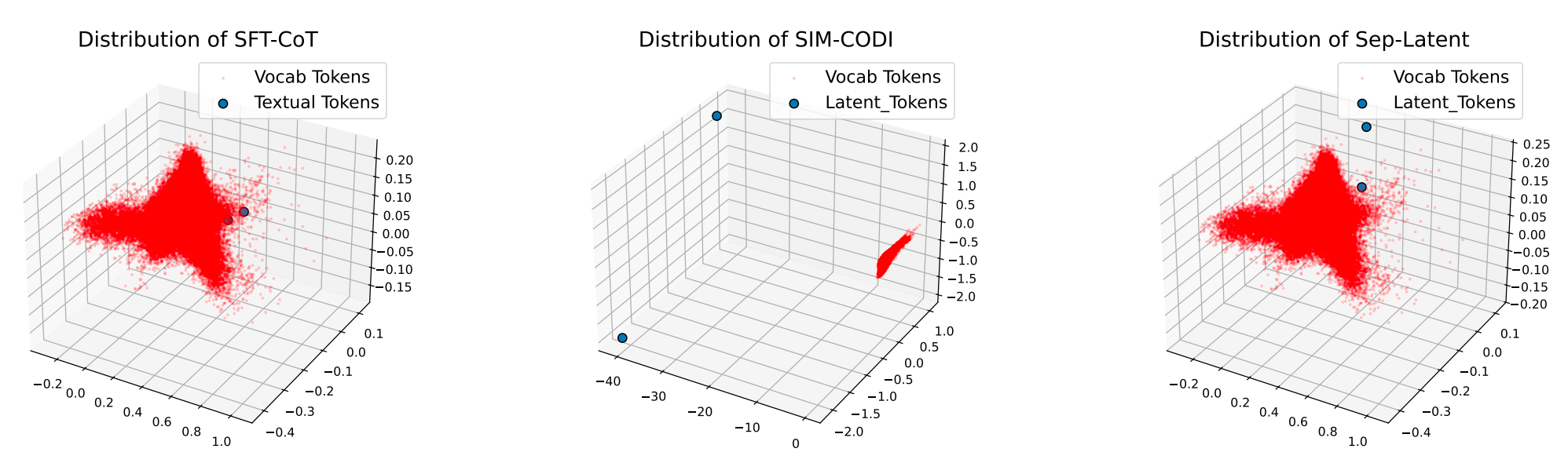}
    \caption{PCA results of token representations across different methods. Sep-Latent stays closer to the vocabulary embedding space than existing implicit reasoning methods.}
    \label{pca-results}
\end{figure*}

\subsection{Efficiency}

\begin{figure}[ht]
    \centering
    \includegraphics[width=.9\columnwidth]{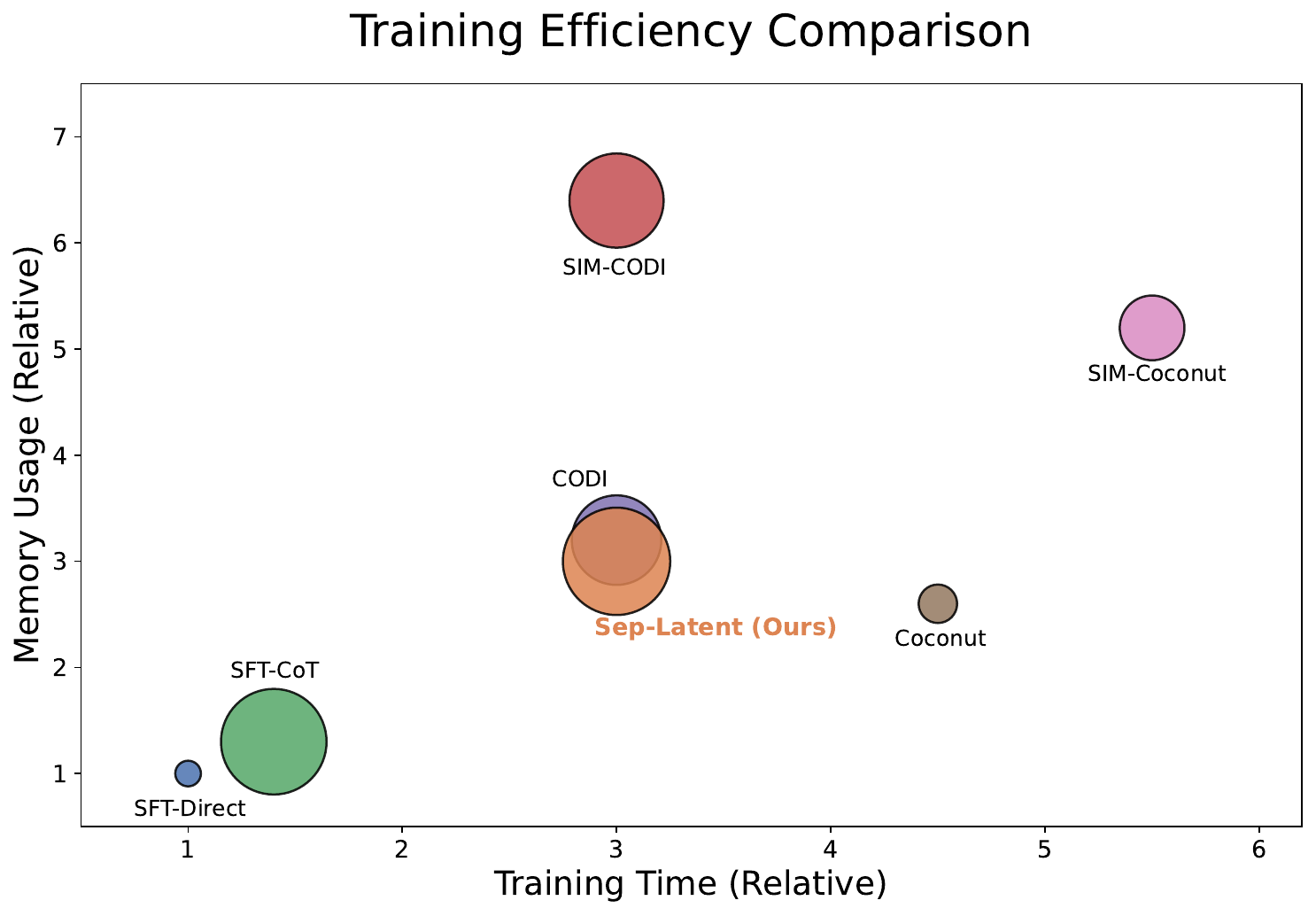}
    \caption{Comparative Analysis of Training Overhead: GPU Memory vs. Training Time
Marker size is proportional to model accuracy (ACC), with larger bubbles representing better performance}
    \label{efficiency}
\end{figure}



As shown in Figure~\ref{efficiency}, we compare methods in training time, GPU memory usage, and accuracy (bubble size). Sep-Latent consistently lies in the best efficiency region, achieving competitive accuracy (56.8\%) with substantially lower training cost than implicit reasoning methods such as Coconut and CODI.

Sep-Latent is also highly token-efficient at inference, requiring only 5.4--5.7 tokens on average, compared with 11.9--13.4 for existing implicit methods and over 23 for SFT-CoT. This efficiency comes from our adaptive reasoning mechanism, where the model learns to emit a termination marker through $\mathcal{L}_{imp\_ce}$ and $\mathcal{L}_{dist\_sep}$, reducing redundant computation while maintaining strong accuracy.

\section{Ablation Studies}

\subsection{Scaling to Larger Backbone}
To evaluate the scalability of our framework, we extend the evaluation to larger models within the Llama3 family: Llama3.2-3B-Instruct and Llama3.1-8B-Instruct.
The results shown in Table~\ref{big-llm} indicate Sep-Latent achieves an in-domain accuracy of 64.7\% on the Llama3-3B, significantly outperforming CODI (60.8\%) and SIM-CODI (62.3\%) while maintaining superior token efficiency with only 5.5 tokens. This trend continues with the Llama3-8B, where our method reaches 68.8\% in-domain accuracy. Notably, across all model scales, Sep-Latent maintains a nearly constant and remarkably low token consumption (5.5–5.7 tokens), whereas other implicit methods require significantly more overhead. These findings demonstrate that our adaptive latent reasoning mechanism retains its efficiency advantages even as the underlying backbone capacity scales.

\subsection{Necessity of Attention-Decay}

\begin{figure}[ht]
    \centering
    \includegraphics[width=.9\columnwidth]{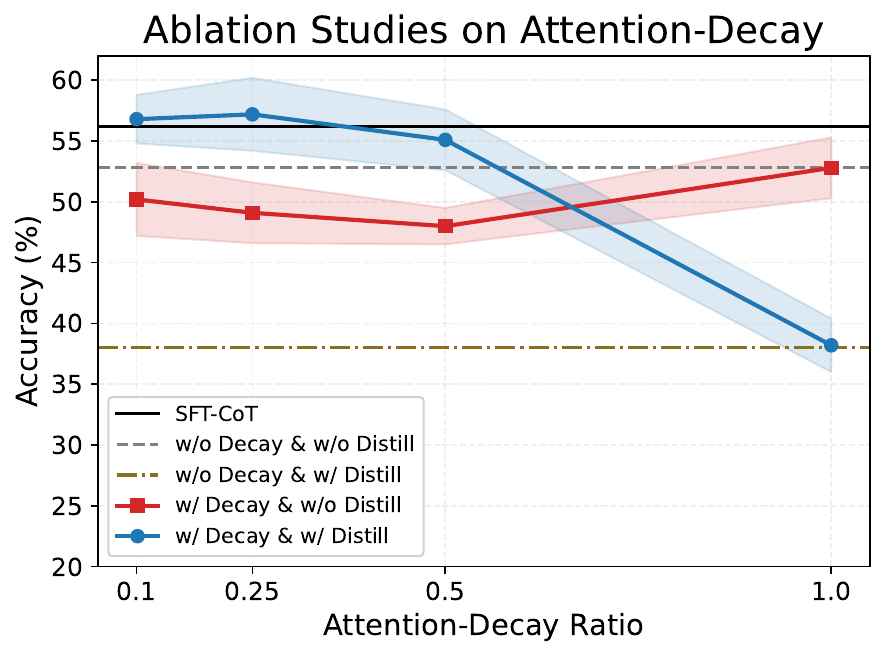}
    \caption{The ablation studies about Attention-Decay Mechanism}
    \label{ablation-decay}
\end{figure}



We evaluated the Attention-Decay mechanism and separator token distillation by repeating each experiment 10 times using top-5 sampling (Figure~\ref{ablation-decay}). These quantitative results are also summarized in Table~\ref{1b-result}. The significant performance drop when applying distillation without decay (green line) suggests that, lacking an explicit attention constraint, separator representations fail to capture the underlying reasoning process, introducing noise rather than guidance. Conversely, implementing decay without distillation (red line) reconfigures attention flow but lacks explicit process-based supervision, which limits its impact on accuracy. The synergy in our full method confirms that Attention-Decay provides the essential structural foundation for separator distillation to meaningfully represent and transfer reasoning logic.



\subsection{Distributions of Latent Tokens}

Recent Studies point out the implicit reasoning~(Coconut) is not really thinking~\cite{zhang2025latent}, it just generates isolated "pause" tokens. We therefore conducted a similar experiment.

The PCA visualization in Figure~\ref{pca-results} shows that Sep-Latent tokens maintain high-quality representations rather than collapsing into uninformative latent states. Unlike drifting implicit representations, our latent tokens remain close to the vocabulary manifold, indicating strong semantic grounding. Moreover, the clearly separated token distributions suggest that different latent tokens encode distinct reasoning computations, forming a semantically structured reasoning space.

\section{Conclusions}


In this paper, we introduced Sep-Latent, a framework for adaptive internalized reasoning that achieves high logical rigor with extreme token efficiency. By utilizing Attention-Decay and separator tokens as semantic anchors, we successfully compress explicit reasoning chains into a structured latent space. Our dual-path distillation ensures that this implicit pathway maintains accuracy without the overhead of textual generation.

Crucially, Sep-Latent enables dynamic reasoning depth, allowing the model to autonomously adjust its latent tokens. This reduces consumption to an average of 5.4 tokens on GSM8k while remaining robust against representational collapse. Our approach represents a significant step toward developing efficient, autonomous reasoning agents that scale effectively across architectures.

\section*{Limitations}
Despite its efficiency and robustness, Sep-Latent has several limitations. First, while our method narrows the performance gap, it does not yet fully surpass the explicit SFT-CoT baseline in accuracy. This suggests that textual chains still provide a more stable supervision signal for certain complex logical tasks that are difficult to capture entirely in the latent space. Second, our sampling analysis reveals that the model often contains the correct answer within its output distribution even when greedy decoding fails. This indicates that the reasoning logic is present, but the model could benefit from Reinforcement Learning (RL) to better align the latent path with the correct solution. 
\bibliography{custom}

\appendix
\section{Inference Mode.} 
\label{inference-mode}
During the inference phase, the explicit CoT path and the attention-decay mechanism are deactivated to maximize computational efficiency. The model generates the final answer by adaptively transitioning from the continuous latent space back to the discrete vocabulary space.

\textbf{Implicit Phase.} Starting from the initial token $v_0$ (e.g., <think>), the model enters the implicit reasoning phase. At each step $t$, the input vector $u_t$ is generated via equation~\ref{implicit-reason}, This allows the model to perform multi-step internal computation within the continuous hidden space, bypassing the overhead of generating intermediate textual tokens.

\textbf{Dynamic Transition.} Unlike existing methods that utilize a fixed number of latent steps, our framework employs an adaptive termination mechanism. At each step $t$ of the implicit generation, the language modeling head $W$ is applied to the current hidden state $h_t^{(L)}$. The implicit phase concludes when the decoded token corresponds to the terminal separator $v_{end}$ (e.g., </think>):
\begin{equation}
    \text{argmax}(W h_t^{(L)}) = \text{index}(v_{end})
\end{equation}
Once this termination condition is met, the model stops the latent projection and reverts to standard autoregressive decoding to generate the final answer $a$. This dynamic transition allows the model to adaptively determine the necessary reasoning depth based on the query complexity, ensuring both logical consistency and superior inference speed.

\section{Datasets}
\subsection{Examples}
An example about our training data in both explict reasoning and implicit reasoning:
\begin{figure}[h]
    \centering
    \begin{tcolorbox}[
        title={Format of Explicit Resoning and Implicit Resoning Data},
    ]
        \begin{minipage}[t]{\linewidth}
        \textbf{Question:}Janet's ducks lay 16 eggs per day. She eats three for breakfast every morning and bakes muffins for her friends every day with four. She sells the remainder at the farmers' market daily for \$2 per fresh duck egg. How much in dollars does she make every day at the farmers' market?
        
        \textbf{Explicit Answer:} <think> 16-3-4=9 <sep> 9*2=18 <sep> </think> 18
        
        \textbf{Implicit Answer:}<think> <latent\_token> <latent\_token> </think> 18
        \end{minipage}
    \end{tcolorbox}
    \label{}
\end{figure}

The <latent\_token> just serves as a placeholder and is not involved in the process of computation.

\label{data-example}
\subsection{Benchmark Distribution}
\label{benchmark-detail}
\begin{figure}[h]
    \centering
    \includegraphics[width=\columnwidth]{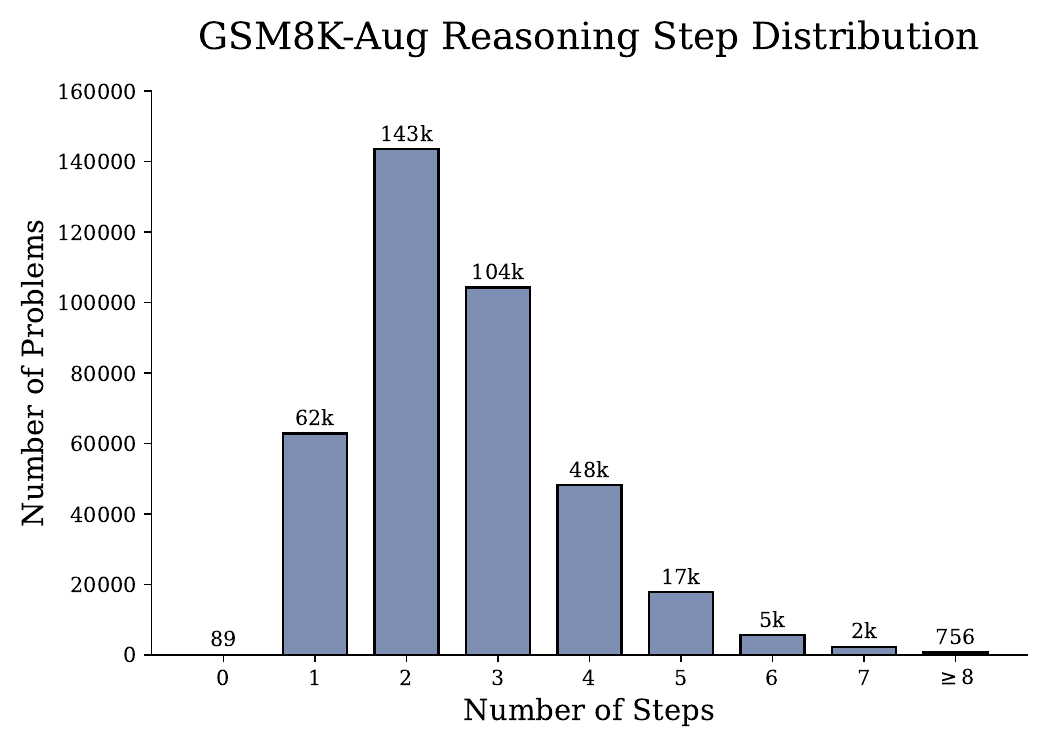}
    \caption{The reasoning steps distribution of training data. Most problems involve 1-4 steps and all the data are used for our training.}
    \label{attention_result_appendix}
\end{figure}

\textbf{GSM-Aug} is the training dataset we use, which is an augmented variant of GSM8K. It expands the original 8.5k training problems to approximately 385k examples through techniques such as paraphrasing, numerical resampling, and synthetic generation using GPT-4. The reasoning steps distribution of it are shown in Figure~\ref{attention_result_appendix}.

\textbf{SVAMP} is an out-of-domain benchmark designed to test the model's structural robustness. It consists of elementary-level math word problems that introduce variations in question phrasing and problem structure to challenge the model's reliance on superficial patterns.

\textbf{GSM-Hard} serves as a challenging out-of-domain evaluation set where numerical values in GSM8k problems are replaced with significantly larger magnitudes. This dataset specifically assesses the model’s ability to generalize its mathematical reasoning capabilities to more complex arithmetic scenarios.

\textbf{GSM-SYM} is a specialized out-of-domain benchmark that introduces symbolic perturbations and variants to the original problems. It is employed to test the stability of reasoning models against subtle linguistic or symbolic changes that do not alter the underlying logic.

\textbf{LogiQA} is a dataset for complex logical reasoning, derived from multiple-choice questions in the National Civil Service Examination. It evaluates the model's capacity for high-level deductive and inductive reasoning.

\textbf{CommonsenseQA (CSQA)} is a large-scale dataset used to assess general world knowledge and commonsense reasoning. It requires models to choose the correct answer by leveraging latent common sense information beyond the provided text.

\textbf{StrategyQA} focuses on implicit reasoning tasks where the required reasoning path is not explicitly stated. Models must infer the correct multi-step strategy and execute the corresponding logical jumps to answer the questions accurately.

\begin{figure*}[htbp]
    \centering
    \includegraphics[width=\linewidth]{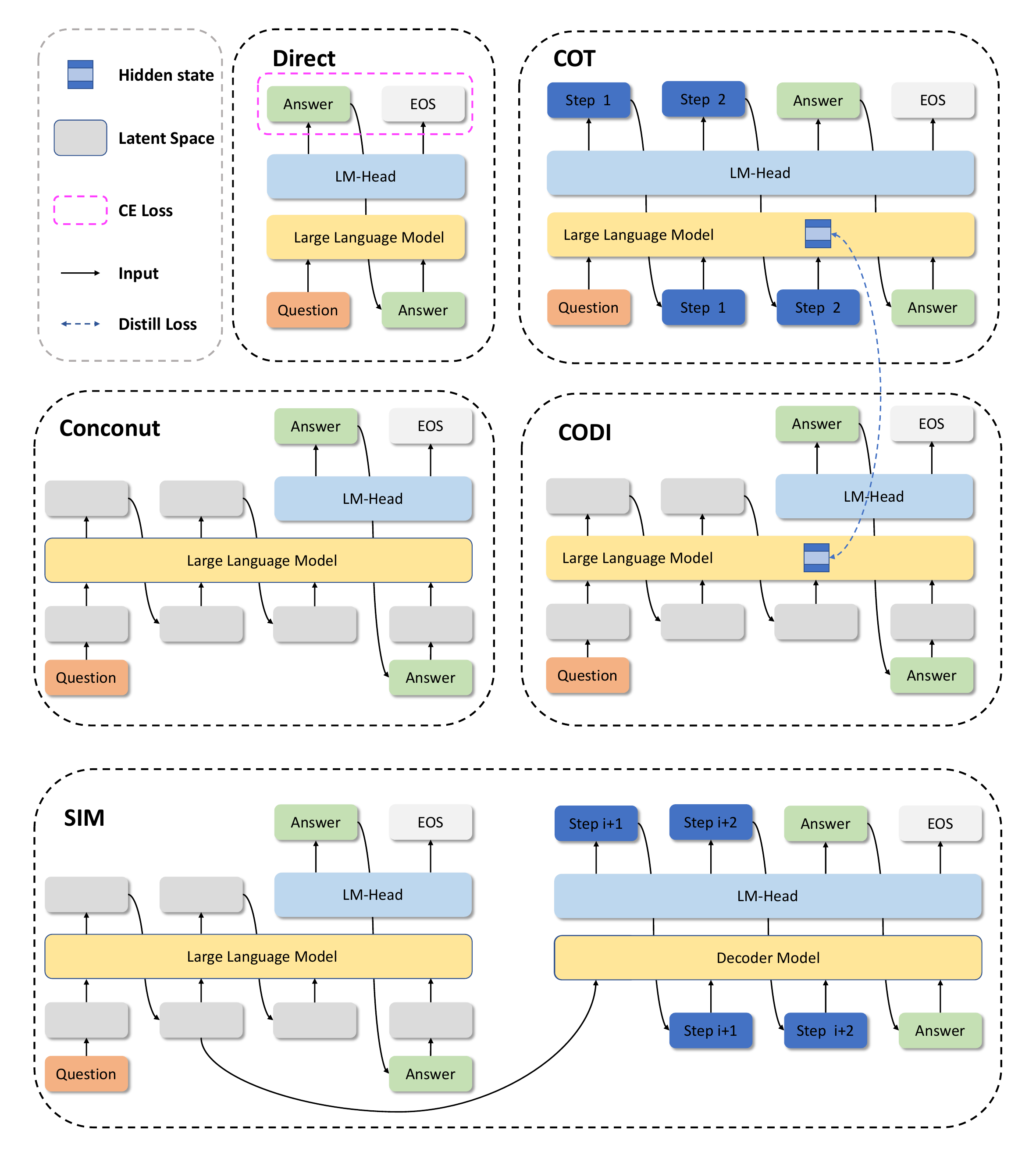}
    \caption{The baselines used in our paper.}
    \label{baselines}
\end{figure*}

\section{implementation Details}
\label{implementation-details}
We implement Sep-Latent using the Hugging Face transformers library and accelerate training via DeepSpeed ZeRO-3. All experiments are conducted on a cluster of H20 GPUs. We fine-tune the model for 3 epochs with a total batch size of 128 and an initial learning rate of 1e-4. The learning rate is governed by a linear warmup followed by a cosine annealing scheduler to ensure stable convergence.
We choose "\textbackslash n" as our <sep> token.

A key component of our training is the curriculum scheduling of the Attention-Decay factor ($\gamma$). To facilitate a smooth transition from standard explicit reasoning to the compressed semantic anchor phase, we employ a dynamic decay strategy during the first epoch. Specifically, $\gamma$ starts at 1.0 (standard attention) and follows a cosine decay schedule until it reaches a predefined target value (e.g., 0.1) by the end of the first epoch. For the remaining two epochs, $\gamma$ is maintained at this constant target value. This curriculum ensures that the model first learns the fundamental logical mappings before being forced to aggregate information into the separator tokens, thereby stabilizing the multi-level distillation process ($\mathcal{L}_{dist\_sep}$).

\section{Difference from Baselines}

\begin{table}[ht]
    \centering
    \begin{tabular}{cccc}
        \toprule
         Model & Method & Lr & Epochs\\
         \midrule
         Llama 1B & SFT-CoT & $1 \times 10^{-4}$ & 3 \\
         Llama 1B & SFT-Direct & $1 \times 10^{-4}$ & 3 \\
         Llama 1B & Coconut & $1 \times 10^{-4}$ & 30 \\
         Llama 1B & SIM-Coconut & $1 \times 10^{-4}$ & 30 \\
         Llama 1B & CODI & $8 \times 10^{-4}$ & 10 \\
         Llama 1B & SIM-CODI & $8 \times 10^{-4}$ & 10 \\
         \bottomrule
    \end{tabular}
    \caption{Training hyperparameters across different baselines.}
    \label{baseline-setting}
\end{table}

To provide a comprehensive comparison, we evaluate our proposed Sep-Latent method against the following representative baselines in Figure~\ref{baselines}:

\textbf{SFT-CoT \& SFT-Direct:} SFT-CoT represents the standard explicit reasoning approach, where the model is fine-tuned to generate a complete textual Chain-of-Thought before the final answer. Conversely, SFT-Direct (or No-CoT) supervises the model to output the answer directly, bypassing any intermediate reasoning process.

\textbf{Coconut:} A curriculum-based framework where the model gradually learns to replace discrete textual reasoning steps with continuous latent tokens over several training stages.

\textbf{CODI:} This method employs iterative self-distillation to align the implicit latent trajectories of a student model with the explicit reasoning traces of a teacher model in the continuous space.

\textbf{SIM-Enhanced Baselines:} We include SIM-Coconut and SIM-CODI to evaluate our method against the current state-of-the-art supervision techniques. These variants integrate the SIM-CoT framework, which utilizes an auxiliary model to decode and supervise the intermediate latent reasoning process, thereby improving logical consistency.

And the training hyperparameters details are shown in Table~\ref{baseline-setting}.

\end{document}